# Preserving Privacy in Surgical Video Analysis Using Artificial Intelligence: A Deep Learning Classifier to Identify Out-of-Body Scenes in Endoscopic Videos

Joël L. Lavanchy, MD[1,2*]; Armine Vardazaryan, MSc[1,3*]; Pietro Mascagni, MD PhD[1,4];

AI4SafeChole Consortium†; Didier Mutter, MD PhD[1,5]; Nicolas Padoy, PhD[1,3]

1 IHU Strasbourg, France
2 Department of Visceral Surgery and Medicine, Inselspital, Bern University Hospital, University of Bern, Switzerland
3 ICube, University of Strasbourg, CNRS, France
4 Fondazione Policlinico Universitario Agostino Gemelli IRCCS, Rome, Italy
5 University Hospital of Strasbourg, France
* Joël Lavanchy and Armine Vardazaryan contributed equally and share first co-authorship
† The AI4SafeChole consortium is represented by Giovanni Guglielmo Laracca, Ludovica Guerriero, Andrea Spota, Claudio Fiorillo, Giuseppe Quero, Segio Alfieri, Ludovica Baldari, Elisa Cassinotti, Luigi Boni, Diego Cuccurullo, Guido Costamagna, and Bernard Dallemagne

This study presents the development, internal and external validation of a deep learning model for the identification of out-of-body images in endoscopic videos. The model revealed a ROC AUC of 99.97%, 99.94%, and 99.71% on an internal and two external datasets, respectively. Out-of-body images can be reliably identified in endoscopic videos to preserve patient's privacy.

**Keywords:** Privacy preservation, out-of-body scenes, endoscopic videos, video-based assessment, artificial intelligence, deep learning, surgical data science.

Correspondence to: Joël L. Lavanchy, MD, joel.lavanchy@ihu-strasbourg.eu

# Index

# Abstract

**Objective:** To develop and validate a deep learning model for the identification of out-of-body images in endoscopic videos.

**Background:** Surgical video analysis facilitates education and research. However, video recordings of endoscopic surgeries can contain privacy-sensitive information, especially if out-of-body scenes are recorded. Therefore, identification of out-of-body scenes in endoscopic videos is of major importance to preserve the privacy of patients and operating room staff.

**Methods:** A deep learning model was trained and evaluated on an internal dataset of 12 different types of laparoscopic and robotic surgeries. External validation was performed on two independent multicentric test datasets of laparoscopic gastric bypass and cholecystectomy surgeries. All images extracted from the video datasets were annotated as inside or out-of-body. Model performance was evaluated compared to human ground truth annotations measuring the receiver operating characteristic area under the curve (ROC AUC).

**Results:** The internal dataset consisting of 356,267 images from 48 videos and the two multicentric test datasets consisting of 54,385 and 58,349 images from 10 and 20 videos, respectively, were annotated. Compared to ground truth annotations, the model identified out-of-body images with 99.97% ROC AUC on the internal test dataset. Mean ± standard deviation ROC AUC on the multicentric gastric bypass dataset was 99.94±0.07% and 99.71±0.40% on the multicentric cholecystectomy dataset, respectively.

**Conclusion:** The proposed deep learning model can reliably identify out-of-body images in endoscopic videos. The trained model is publicly shared. This facilitates privacy preservation in surgical video analysis.

## 1. Introduction

Surgical video analysis facilitates education (review of critical situations and individualized feedback),[1,2] credentialing (video-based assessment)[3] and research (standardization of surgical technique in multicenter trials,[4] surgical skill assessment).[5,6] Despite its increasing use, the full potential of surgical video analysis has not been leveraged so far, as manual case review is time-consuming, costly, needs expert knowledge and raises privacy concerns.

Therefore, surgical data science approaches have been adopted recently to automate surgical video analysis. Artificial intelligence (AI) models have been trained to recognize phases of an intervention,[7-9] tools[7,10] and actions[11] in surgical videos. This allows for down-stream applications like the estimation of the remaining surgery duration,[12] automated documentation of critical events,[13] assessment of surgical skill[14] and safety check-point achievement,[15] and intraoperative guidance.[16]

AI will continue to reduce the costs and time constraints of experts reviewing surgical videos. However, the privacy concerns regarding the handling of patient video data have not been extensively addressed so far. Endoscopic videos often contain scenes of the operating room (OR) that could potentially reveal sensitive information such as the identity of patients or OR staff. Moreover, if clocks or calendars present in the room are captured in the video, the time or date of the respective intervention can be identified. These scenes recorded outside of the patient's body are referred to as out-of-body scenes. If video recording has already been started before the endoscope is introduced into the patient, has not been stopped after the surgery was terminated or every time the endoscope is cleaned during surgery, out-of-body scenes are captured.

This article reports the development and validation of a deep learning-based image classifier to identify out-of-body scenes in endoscopic videos, called Out-of-Body Network (OoBNet). External validation of OoBNet is performed on two independent multicentric datasets of laparoscopic Roux-en-Y gastric bypass and laparoscopic cholecystectomy surgeries.

Deep learning is a type of machine learning using artificial neural networks. Within a deep neural network, the input data is processed in multiple layers of neurons leading to abstract representations of the data. This allows the model to predict whether a given input frame (image) is inside the body or out-of-body. Based on frame-wise classification, OoBNet enables privacy protection of patients and OR staff through automated recognition of out-of-body scenes in endoscopic videos. Once identified, out-of-body scenes can easily be blurred or deleted at the discretion of the respective use-case. The trained model and an executable application of OoBNet are published, to provide an easy-to-use tool for surgeons, data scientists and hospital administrative staff to anonymize endoscopic videos.

## 2. Methods

### 2.1 Datasets

The dataset used for the development of OoBNet was created from surgeries recorded at the University Hospital of Strasbourg, France.[17] Four video recordings for each of the following endoscopic procedures were arbitrarily selected: Laparoscopic Nissen fundoplication, Roux-en-Y gastric bypass, sleeve gastrectomy, hepatic surgery, pancreatic surgery, cholecystectomy, sigmoidectomy, eventration, adrenalectomy, hernia surgery, robotic Roux-en-Y gastric bypass, and robotic sleeve gastrectomy. The dataset containing 48 videos was split into training-, validation-, and test-set including 2, 1 and 1 videos of each procedure, respectively.

External validation of the model was done on a random sample of 5 videos from 6 centers and two independent multicentric datasets. 1) A dataset of 140 laparoscopic Roux-en-Y gastric bypass videos from the University Hospital of Strasbourg, France and Inselspital, Bern University Hospital,

Switzerland.[18] 2) A dataset of 174 laparoscopic cholecystectomy videos from four Italian centers: Policlinico Universitario Agostino Gemelli, Rome; Azienda Ospedaliero-Universitaria Sant'Andrea, Rome; Fondazione IRCCS Ca' Granda Ospedale Maggiore Policlinico, Milan; and Monaldi Hospital, Naples. This dataset was collected for the multicentric validation of EndoDigest, a computer vision platform for video documentation of the critical view of safety (CVS).[19]

An illustration of the dataset split for model development, internal and multicentric external validation is displayed in **Figure 1**.

Each hospital complied with local institutional review board requirements. Patients either consented to the recording of their intervention or to the use of their health record for research purposes. All videos were shared as raw video material without identifying metadata.

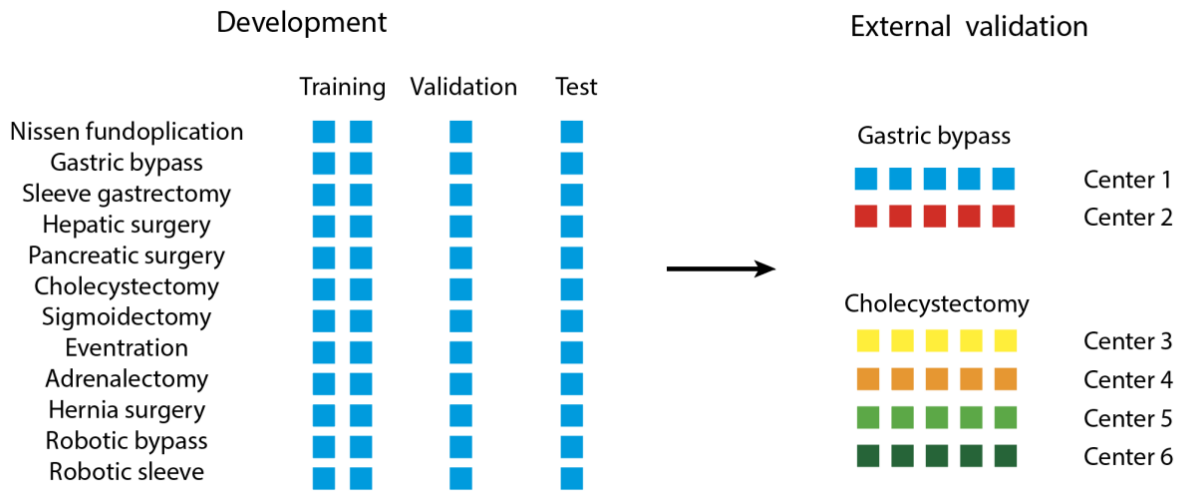

**Figure 1. Illustration of dataset splits for model development, internal and external validation.** Every square represents a video. Videos from the same center have the same color.

## 2.2 Annotations

Each video was split into frames at the rate of 1 frame per second. All frames were annotated in a binary fashion, being either inside the patient's abdomen or out-of-body. The valve of the trocar was the visual cue for the transition from inside to out-of-body. All frames in which the valve of the optic trocar is visible are considered out-of-body to err on the safe side of privacy preservation. All datasets were annotated by a single annotator (A.V.). Edge cases were reviewed by a board-certified surgeon with extensive experience in surgical video analysis (J.L.L.).

## 2.3 Model architecture and training

OoBNet is a deep learning-based image classifier, using MobileNetV2[20] as backbone followed by dropout (with dropout rate 0.5), a long short-term memory (LSTM with 640 units)[21], linear and sigmoid layers. Layer normalization was applied before dropout and linear layers. MobileNetV2 is a model architecture designed for image recognition at low computational resources as in mobile devices and smartphones. The LSTM layer contains memory gates that bring context awareness to frame classification. As part of preprocessing, input images were resized to 64x64 pixels, then augmented with random rotation and contrast. Data augmentation is a common way to generate variance in the input dataset to improve the robustness of the model. The output of OoBNet is a probability-like value that is then binarized to either 0 or 1 to predict whether the image is an inside or out-of-body frame (**Figure 2**).

The network was trained on video clips of 2,048 consecutive frames for 300 epochs (cycles) with early stopping applied according to the highest F1-score obtained on the validation dataset. The optimizer used was Adam[22] with a learning rate of 0.00009 and a batch size of 2,048. The trained model and an executable application of OoBNet are available at https://github.com/CAMMA-public/out-of-body-detector.

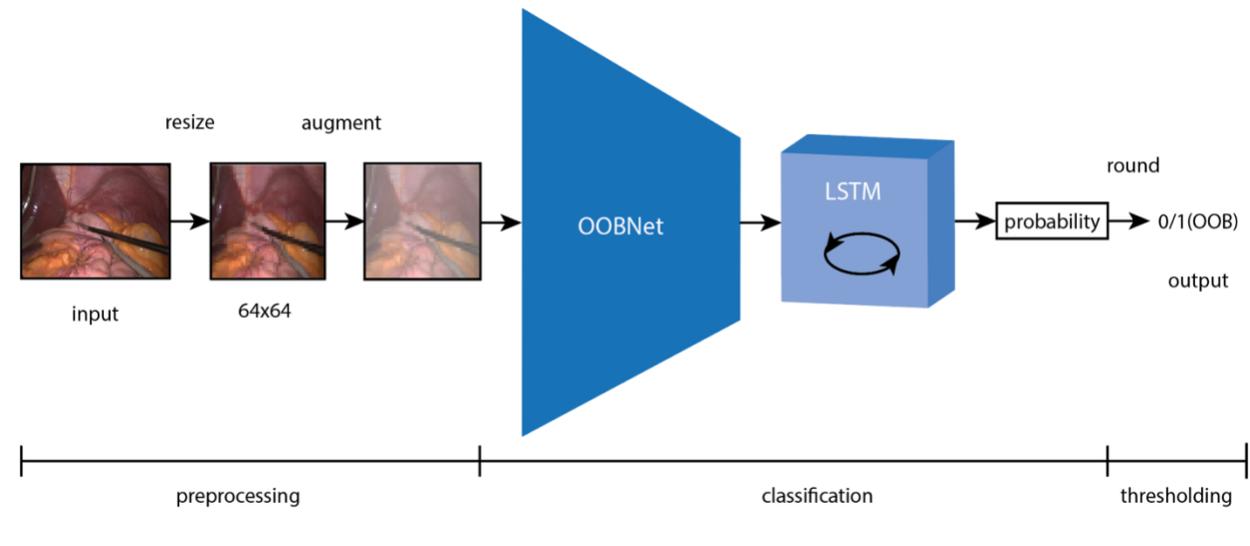

**Figure 2. Model architecture of OoBNet.** The input image is resized to 64x64 pixels and augmented with random rotation and contrast. Then it is fed to the deep neural network with a consecutive long short-term memory (LSTM) which outputs a probability like value whether the image is out-of-body or not. This probability is rounded at a 0.5 threshold to either 0 (inside the body) or 1 (out-of-body OOB).

## 2.4 Model evaluation

OoBNet was evaluated on the test dataset, which was used neither for model training nor validation. Furthermore, external evaluation was done on two independent and multicentric datasets as described above. The predictions of OoBNet were compared to human ground truth annotations. The performance of OoBNet was measured as precision, recall, F1-score, average precision, and receiver operating characteristic area under the curve (ROC AUC). Precision is the proportion of true positives among all positive predictions (true and false positives), also referred to as positive predictive value. Recall is the proportion of true positives among all relevant predictions (true positives and false negatives), also referred to as sensitivity. F1-score is the harmonic mean of precision and recall. Average precision is the area under the precision-recall curve. ROC AUC is the area under the receiver operating characteristic curve which is created by plotting sensitivity against 1-specificity. It is also referred to as c-statistic.

## 3. Results

OoBNet was trained, validated, and tested on an internal dataset of 48 videos with a mean duration ± standard deviation (**SD**) of 123±79 minutes containing a total of 356,267 frames. Thereof, 112,254 (31.51%) were out-of-body frames. External validation of OoBNet was performed on a gastric bypass dataset of 10 videos with a mean duration ± SD of 90±27 minutes containing a total of 54,385 frames (4.15% out-of-body frames) and on a cholecystectomy dataset of 20 videos with a mean duration ± SD of 48±22 minutes containing a total of 58,349 frames (8.65% out-of-body frames). The full dataset statistics and the distribution of frames across training, validation and test set are displayed in **Table 1.**

The ROC AUC of OoBNet evaluated on the test set was 99.97%. Mean ROC AUC ± SD of OoBNet evaluated on the multicentric gastric bypass dataset was 99.94±0.07%. Mean ROC AUC ± SD of OoBNet evaluated on the multicentric cholecystectomy dataset was 99.71±0.40%. The full quantitative results are shown in **Table 2**. Confusion matrices on the test set, the multicentric gastric bypass dataset, and the multicentric cholecystectomy dataset are displayed in **Figure 3A-G**. OoBNet was evaluated on a total of 111,974 frames, whereof 557 frames (0.50%) were falsely classified as inside the body even though they were out-of-body frames (false negative prediction). Qualitative results illustrating false positive and false negative predictions of OoBNet are displayed in **Figure 4.**

**Table 1: Dataset statistics**

| Dataset | Videos | Min. duration | Max. duration | Average duration | Total frames | Out-of-body frames | |
|---|---|---|---|---|---|---|---|
| | (n) | (min.) | (min.) | (min.) | (n) | (n) | (%) |
| Training | 24 | 35 | 302 | 112 | 161,870 | 57,203 | 35.34 |
| Validation | 12 | 26 | 408 | 132 | 95,157 | 23,388 | 24.58 |
| Test | 12 | 33 | 239 | 137 | 99,240 | 31,663 | 31.91 |
| Gastric bypass | 10 | 55 | 134 | 90 | 54,385 | 2,259 | 4.15 |
| Center 1 | 5 | 66 | 134 | 101 | 30,464 | 696 | 2.28 |
| Center 2 | 5 | 55 | 117 | 79 | 23,921 | 1,563 | 6.53 |
| Cholecystectomy | 20 | 17 | 88 | 48 | 58,349 | 5,050 | 8.65 |
| Center 3 | 5 | 31 | 88 | 55 | 16,786 | 2,268 | 13.51 |
| Center 4 | 5 | 24 | 88 | 46 | 13,997 | 885 | 6.32 |
| Center 5 | 5 | 44 | 76 | 57 | 17,371 | 356 | 2.05 |
| Center 6 | 5 | 17 | 53 | 34 | 10,195 | 1,541 | 15.12 |

**Table 2: Quantitative evaluation results on internal and external datasets**

| Dataset | Videos (n) | ROC AUC (%) | Average precision (%) | F1-score (%) | Precision (%) | Recall (%) |
|---|---|---|---|---|---|---|
| Test | 12 | 99.97 | 99.94 | 99.50 | 99.69 | 99.31 |
| *Gastric bypass | 10 | 99.94±0.07 | 99.42±0.26 | 96.10±1.53 | 98.39±0.81 | 93.96±3.67 |
| Center 1 | 5 | 99.99 | 99.60 | 97.18 | 97.82 | 96.55 |
| Center 2 | 5 | 99.89 | 99.23 | 95.01 | 98.96 | 91.36 |
| *Cholecystectomy | 20 | 99.71±0.40 | 98.66±1.27 | 94.74±2.17 | 95.43±5.62 | 94.37±4.01 |
| Center 3 | 5 | 99.83 | 99.00 | 92.78 | 87.20 | 99.12 |
| Center 4 | 5 | 99.92 | 98.93 | 96.27 | 97.79 | 94.80 |
| Center 5 | 5 | 99.12 | 96.85 | 92.98 | 96.95 | 89.33 |
| Center 6 | 5 | 99.97 | 99.85 | 96.93 | 99.79 | 94.22 |

*Results are mean ± standard deviation; ROC AUC, receiver operating characteristic area under the curve

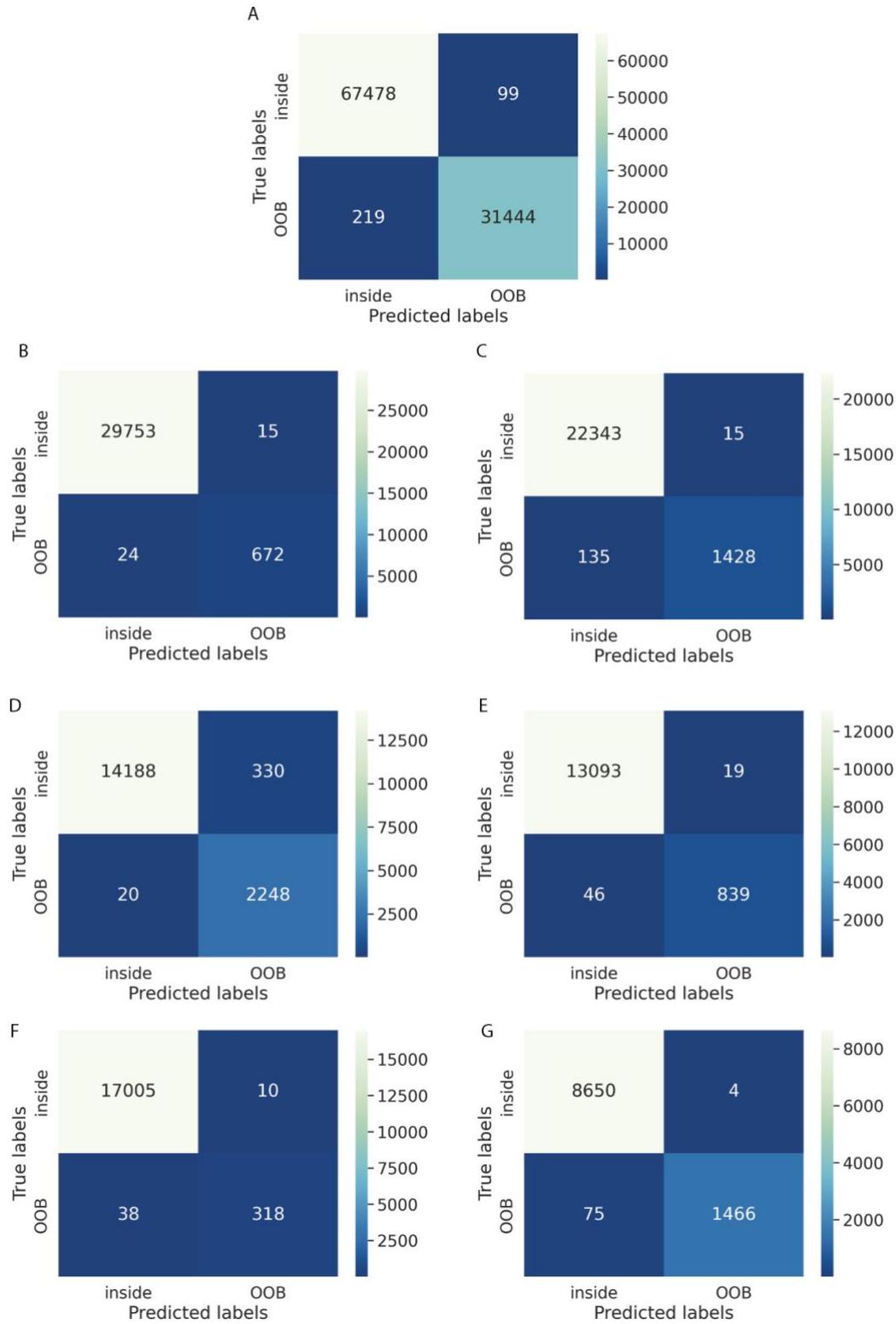

**Figure 3. Confusion matrices.** A: test set; B and C: centers 1 and 2 (multicentric gastric bypass dataset); D-G: centers 3, 4, 5, and 6 (multicentric cholecystectomy dataset).

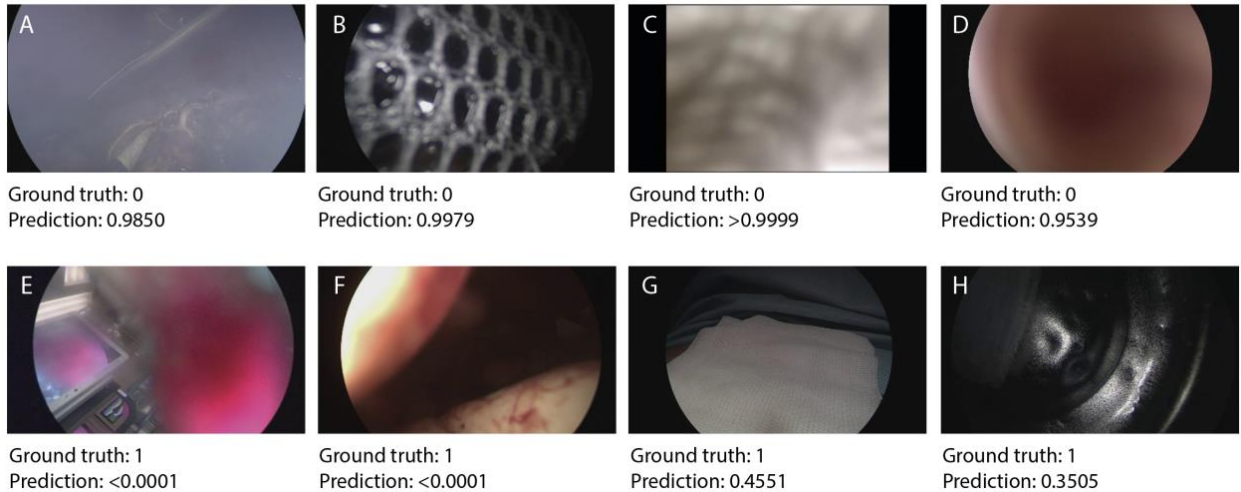

**Figure 4: Qualitative results.** Top row: False positive model predictions (OoBNet predicts the frame to be out-of-body even though it is not). Bottom row: False negative model predictions (OoBNet predicts the frame to be inside the body even though it is out-of-body). Below each image the binary human ground truth annotations and the probability like model predictions are provided. In A, surgical smoke is impairing the vision. In B, C, and D, a mesh, a swab, and tissue are so close, that – lacking the temporal context – it is difficult to distinguish even for a human annotator whether it is out-of-body or not. In E and F, blood on the endoscope and a glove with bloodstains mimic an inside view. In G, a surgical towel covers most of the patient's body, so that the model lacks visual cues for an out-of-body frame. In H, the endoscope is cleaned in a thermos, which mimics the inside of a metal trocar.

## 4. Discussion

This study reports the development and multicentric validation of a deep learning-based image classifier to detect out-of-body frames in endoscopic videos. OoBNet showed a performance of 99% ROC AUC in validation on three independent datasets. Using the provided trained model or the executable application, OoBNet can easily be deployed to anonymize endoscopic videos in a retrospective fashion. This allows to create video databases while preserving the patient's and OR staff's privacy and furthermore facilitates the use of endoscopic videos for educational or research purposes without revealing any sensitive information.

To our knowledge, OoBNet is the first out-of-body image classifier trained on videos of multiple interventions and validated on two external datasets. Previous work by our group used an unsupervised computer vision approach to identify out-of-body frames. Based on the redness- and brightness-levels of images, they were classified at an empirically set threshold as inside the body or out-of-body.[23] Another work used a semi-supervised machine learning approach to detect out-of-body scenes in a large dataset of laparoscopic cholecystectomy videos yielding a 97% accuracy.[24] However, this previous study has two major limitations. On the one hand, the main performance metric reported is accuracy. Accuracy is sensitive to the data distribution, or the prevalence of a given observation. On the other hand, it was trained on a dataset of a single intervention type only. This does not ensure that the model generalizes to other intervention types.

Despite OoBNet's excellent performance, even in external validation, not every frame was correctly classified. The ideal classifier would have neither false positive (predicted as out-of-body by the model though inside the body) nor false negative predictions (predicted as inside the body by the model though out-of-body). However, to err on the privacy preserving site, false negative predictions must be minimized. In other words, the threshold of the classifier needs to be optimized for sensitivity (recall). But maximum sensitivity and no false negatives predictions only can be achieved if every frame is classified as out-of-body. Though, this would be a completely unspecific classifier leading to a complete loss of the inside of body frames, which are relevant for surgical video analysis. Therefore, a tradeoff between precision and recall needs to be done. As F1-score is the harmonic mean of precision and recall, a classifier at maximum F1-score optimizes precision and recall at the same time. In this study the maximum F1-score on the validation set was achieved at a classifier threshold of 0.73. But as this threshold yielded slightly more false negative predictions in favor of less false positive predictions, we used the default threshold of 0.5.

As qualitative results show (**Figure 4**), the performance of OoBNet was limited if the endoscopic vision was impaired by surgical smoke, fog, or blood. Furthermore, OoBNet predicted false positive results when objects (mesh, swabs, tissue) were so close to the camera, that the vision was blurred, and even for a human annotator it was difficult to distinguish whether a given frame is out-of-body or not. Moreover, OoBNet predicted false negative results if an out-of-body frame visually resembled an inside scene. Manual inspection of false negative predictions (n=557) on all test datasets revealed three privacy sensitive frames, in which OR staff potentially could have been identified. However, out of 111,974 frames OoBNet was evaluated on not a single frame revealed the identity of the patient, the time, or the date of the intervention.

In external validation OoBNet showed a drop of up to 6.7% points F1-score. This is in line with results from multicentric validation of other AI models in the surgical domain. For example, state of the art surgical phase recognition models have demonstrated variable performance in multicentric validation.[25,26] Furthermore, EndoDigest, a computer vision platform for video documentation of CVS in laparoscopic cholecystectomy, showed a 64-79% successful CVS documentation when validated on a multicentric external dataset compared to 91% successful CVS documentation on the internal dataset.[13,19] Therefore, the performance of AI models trained and evaluated on a single dataset should be regarded cautiously, and these results further highlight the need for external validation of AI models. Our model, however, has shown to generalize well on videos from several external centers.

In conclusion, OoBNet can identify out-of-body frames in endoscopic videos of our datasets with a 99% ROC AUC. It has been extensively validated on internal and external multicentric datasets. OoBNet can be used with high reliability to anonymize endoscopic videos for archiving, research, and education.

# 5. Disclosures

## 5.1 Acknowledgements

Joël Lavanchy was funded by the Swiss National Science Foundation (P500PM_206724). This work was supported by French state funds managed by the ANR within the National AI Chair program under Grant ANR-20-CHIA-0029-01 (Chair AI4ORSafety) and within the Investments for the future program under Grant ANR-10-IAHU-02 (IHU Strasbourg).

## 5.2 Data availability statement

The code of the model and the trained model weights are available at https://github.com/CAMMA-public/out-of-body-detector.

Due to privacy restrictions the datasets used in the present work cannot be publicly shared.